\documentclass[letterpaper,11pt]{article}
\usepackage[utf8]{inputenc}
\usepackage{fancyvrb}
\usepackage{color}
\usepackage[usenames,dvipsnames]{xcolor}
\usepackage{amsmath,amsfonts,amsthm,amssymb}
\usepackage{mathrsfs} 
\usepackage{graphicx}
\usepackage{multirow}
\usepackage{fancyhdr}
\usepackage{extarrows}
\usepackage[margin=1in]{geometry}
\usepackage{framed}
\usepackage{tikz}
\usepackage{float}
\usepackage{subfigure}
\usepackage{algorithm}
\usepackage[noend]{algpseudocode}
\usetikzlibrary{arrows}
\usepackage{tabularx}
\usepackage{hyperref}
\usepackage{fec,dpr}
\usepackage{enumerate}
\usepackage{ragged2e}
\hypersetup{colorlinks}

\newcommand{\algcro}  {\text{HWGANs}}
\newcommand{\alexrnn}  {\text{Alex's Generator}}

\definecolor{mypink}{rgb}{0.858, 0.188, 0.478}

\def \S {\mathbb{S}}

\graphicspath{{./images}}

\makeatletter
\def\BState{\State\hskip-\ALG@thistlm}
\makeatother

\fvset{frame=single,framesep=1mm,fontfamily=courier,fontsize=\scriptsize,numbers=left,framerule=.3mm,numbersep=1mm,commandchars=\\\{\}}

\tikzset{
  treenode/.style = {align=center, inner sep=0pt, text centered,
    font=\sffamily},
  arn_n/.style = {treenode, circle, white, font=\sffamily\bfseries, draw=black,
    fill=black, text width=1.5em},
  arn_r/.style = {treenode, circle, red, draw=red,
    text width=1.5em, very thick},
  arn_x/.style = {treenode, rectangle, draw=black,
    minimum width=0.5em, minimum height=0.5em}
}

\title{Generative Adversarial  Network for Handwritten Text}

\author{Bo Ji\footnotemark[1]\ \ and  Tianyi Chen\footnotemark[2]}

\begin{document}
\maketitle

\renewcommand{\thefootnote}{\fnsymbol{footnote}}
\footnotetext[1]{Zhejiang University, Email: \href{mailto:gsorris27@gmail.com}{gsorris27@gmail.com}}
\footnotetext[2]{Microsoft Cloud \& AI, Email: \href{mailto:Tianyi.Chen@microsoft.com}{Tianyi.Chen@microsoft.com}}
\renewcommand{\thefootnote}{\arabic{footnote}}

\begin{abstract}
Generative adversarial networks (GANs) have proven hugely successful in variety of applications of image processing. However, generative adversarial networks for handwriting is relatively rare somehow because of difficulty of handling sequential handwriting data by Convolutional Neural Network (CNN). In this paper, we propose a handwriting generative adversarial network framework (HWGANs) for synthesising handwritten stroke data. The main features of the new framework include: (i) A discriminator consists of an integrated CNN-Long-Short-Term-Memory (LSTM) based feature extraction with Path Signature Features (PSF) as input and a Feedforward Neural Network (FNN) based binary classifier; (ii) A recurrent latent variable model as generator for synthesizing sequential handwritten data. The numerical experiments show the effectivity of the new model.  Moreover, comparing with sole handwriting generator, the HWGANs synthesize more natural and realistic handwritten text.
\end{abstract}

\section{Introduction}

Learning generative sequential models is a long-standing machine learning challenge and historically in the domain of dynamic Bayesian networks (DBNs) such as Hidden Markov Model (HMMs), then further handled by Recurrent Neural Network. Among various  sequential data, handwritten text has served for thousands of years as one of our primary means of communication. Nowadays, following the rapid development of electronic engineering, human being utilizes more and more stylus-based devices, like smartphone, whiteboard, or tablet to write text as digital ink which are easily processed and manipulated. However, despite promising progress in handwriting recognition, synthesizing handwritten text in digital ink is still relatively less exploited. Even though there exist a few well-designed RNN-based generative models~\cite{graves2013generating, DBLP:journals/corr/abs-1801-08379, DBLP:journals/corr/ChungKDGCB15} for handwritten text, it has been observed that sometimes, the synthesized texts of these models 
are not as natural as human-being's normal written text. 

On the other hand, generative adversarial networks (GANs) are a merging technique proposed in 2014~\cite{goodfellow2014generative} to learn deep representation without numerous annotated training data. In literature, GANs consist of an expert, known as Discriminator $\mathcal{D}$, and a forger, known as Generator $\mathcal{G}$~\cite{creswell2018generative}.  The forger creates forgeries with the aim of faking realistic images, while the expert aims at distinguishing between genuine and faked images. This architecture achieves the generation of high reality, consequently the representation learned by GANs may be used in a variety of applications of image processing, while currently less applied on sequential data generation. Therefore, a common analogy is to think of bringing the idea of GANs into handwritten text generation of digital ink in order to achieve more realistic handwritten text than handwritten generator alone.

In this paper, we propose a generative adversarial architecture to generate realistic handwritten strokes. The following bulleted points summarize our key contributions.
\begin{itemize}
\item We design a discriminator consisting of a compact CNN-LSTM based feature extraction, followed by an auxiliary feedforward neural network to distinguish between genuine and forgery hand-writing. The input of this feature extraction is encoded handwritten strokes following Path Signature Features~\cite{chen2017compact}. 
\item We construct an adversarial architecture by combining the designed discriminator and the handwritten generator proposed by Alex Grave~\cite{graves2013generating}, referred as \alexrnn{} throughout the whole remaining paper. 
\item We use \algcro{} to obtain more realistic English handwritten text than \alexrnn{}. In general, \algcro{} generate handwritten texts which distribute more spatially uniformly than \alexrnn. On the other hand, the handwritten samples synthesized by \algcro{} are typically more neat and sometimes possess more diverse styles than \alexrnn.
\end{itemize}

The remainder of this paper is organized as follows. First, in Section~\ref{sec.relatedwork}, we briefly review some related work about handwritten text generation. Next, in Section~\ref{sec.gan}, we move on to describe our proposed handwritten GANs (\algcro), followed by  numerically demonstrating the effectiveness of \algcro{} in Section~\ref{sec.experiments}. And we finally conclude this paper in Section~\ref{sec.concolusion}. 

\section{Related Work}\label{sec.relatedwork}

\textbf{Handwritten Text Image GANs}: The GANs proposed on 2019~\cite{alonso2019adversarial} aims at generating realistic images of handwritten texts, which is naturally a fit for Optical Character Recognition (OCR). The authors use bidirectional LSTM recurrent layers to get an embedding of the word to be rendered, and then feed it to the generator network. They also modify the standard GAN by adding an auxiliary network for text recognition. However, since this approach can not directly synthesize handwritten text of digital ink, although its generated images are realistic, an additional effective Ink Grab algorithm is further required for the conversion from image to digital stroke.  

\textbf{Variational RNN (VRNN) Handwritten Generator}: VRNN~\cite{chung2015recurrent} explores the inclusion of latent random variables into the hidden state of a RNN by combining a recurrent variational autoencoder (VAE), which is effective modeling paradigm to recover complex multimodal distributions over the non-sequential data space~\cite{kingma2013auto}. This VRNN has been numerically demonstrated by generating speech and handwritten data, while their synthesized handwritten data are not realistic enough.  

\textbf{Alex Grave's Handwritten Generator}: Alex Grave proposed an RNN based generator model to mimic handwriting data~\cite{graves2013generating}, referred as \alexrnn{} throughout the whole paper. For each timestamp, \alexrnn{} encodes prefix sampled path to produce a set of parameters of a probability distribution of next stroke point, then sample the next stroke point given this distribution.  There are two variants of \alexrnn{}, i.e., handwritten predictor and handwritten synthesizer, where the later one has the capability to synthesize handwritten text for given text.


\section{Proposed Adversarial Approach}\label{sec.gan}

In this section, we present our proposed adversarial approach for digital ink handwritten text generation. As standard architectures of GANs, \algcro{} comprise a discriminator $\mathcal{D}$ and a generator $\mathcal{G}$. Discriminator $\mathcal{D}$ is used to distinguish realistic handwritten text from forgery ones, and described in Section~\ref{subsec.discriminator}.  In Section~\ref{subsec.generator},  we summarize generator $\mathcal{G}$, which is designed to forge realistic handwritten text to cheat discriminator adversarially. 

For a more rigorous description, let us introduce the notations used throughout the whole remaining paper. Denote $H=\{S_{1},S_{2}, \cdots, S_{n}\}$ as a handwritten text instance, which consists of a sequence of strokes, where $S_{i}$ refers as the $i$-th stroke of current handwritten instance. One stroke $S$ comprises a sequential stroke points, i.e., $S=\{(x_1,y_1,s_1), (x_2,y_2,s_2), \cdots, (x_m,y_m,s_m)\}$, where $(x_i, y_i, s_i)$ is $i$-th stroke point with $(x_i, y_i)$ as its coordinate axis and $s_i\in \{0,1\}$ as pen up/down status respectively to indicate the starting or ending stroke point.

\subsection{Discriminator}\label{subsec.discriminator}

Discriminator $\mathcal{D}$ is essentially a binary classifier to predict handwritten text of digital ink as a forgery or not. Consequently, feature extraction is important for successful classification. Since the target handwritten data $H$ is sequential, handling the task by recurrent network seems a natural choice. However, digital ink representations of the same handwritten text, especially the number of stroke points, may be varied dramatically, among different underlying ink representation algorithms, and may further cause significant classification discrepancy by  recurrent network. To solve this issue, rendering stroke points into binary images seems a solution, while lots of useful information like order and evolution of stroke points are lost after rendering. In order to balance rendering technique and sequential property of handwritten data, we uniformly sample the stroke points and further utilize Path Signature Features (PSF) which has been demonstrated the effectiveness in online handwritten text recognition~\cite{chen2017compact}. 

\begin{figure}[ht!]
\centering
\includegraphics[scale=0.5]{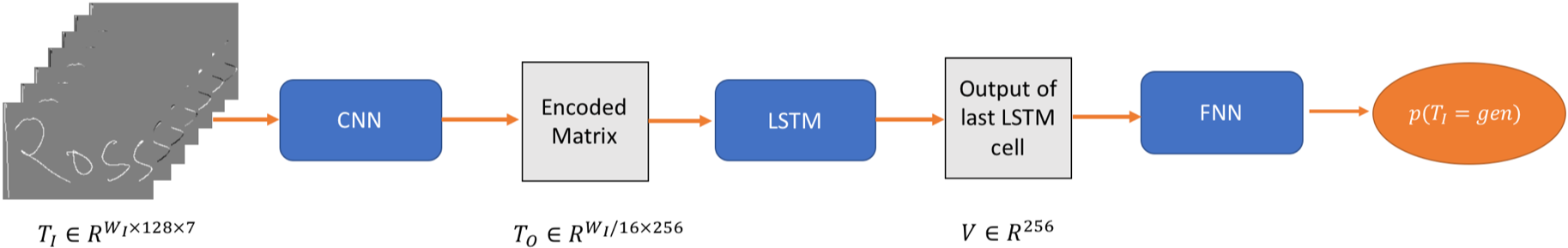}
\caption{Digram of CNN-LSTM model.}
\label{figure:cnn-lstm}
\end{figure}

Path signature feature is defined over a path, and calculated up to 2nd order~\cite{chen2017compact}. More specifically, for each smallest stroke point path $\{(x_i,y_i), (x_{i+1}, y_{i+1})\}$ consisting of 2 consecutive points, the signature feature $p_{i,i+1}=(p_{i,i+1}^{(0)}, p_{i,i+1}^{(1)}, p_{i,i+1}^{(2)})\in\mathbb{R}^7$ is calculated as follows: 
\begin{equation}\label{eq.def:psf}
\begin{split}
p_{i,i+1}^{(0)}&=1\in \mathbb{R} \\
p_{i,i+1}^{(1)}&=(x_{i+1},y_{i+1})-(x_i, y_i)\in \mathbb{R}^2\\
p_{i,i+1}^{(2)}&=p_{i,i+1}^{(1)}\otimes p_{i,i+1}^{(1)}\in \mathbb{R}^4
\end{split}
\end{equation}
where $\otimes$ represents Kronecker matrix product. It follows the definition of PSF~\eqref{eq.def:psf} that PSF encodes geometrical and order information of stroke points into seven path signature maps. 
\begin{figure}[ht!]
\centering
\includegraphics[scale=0.5]{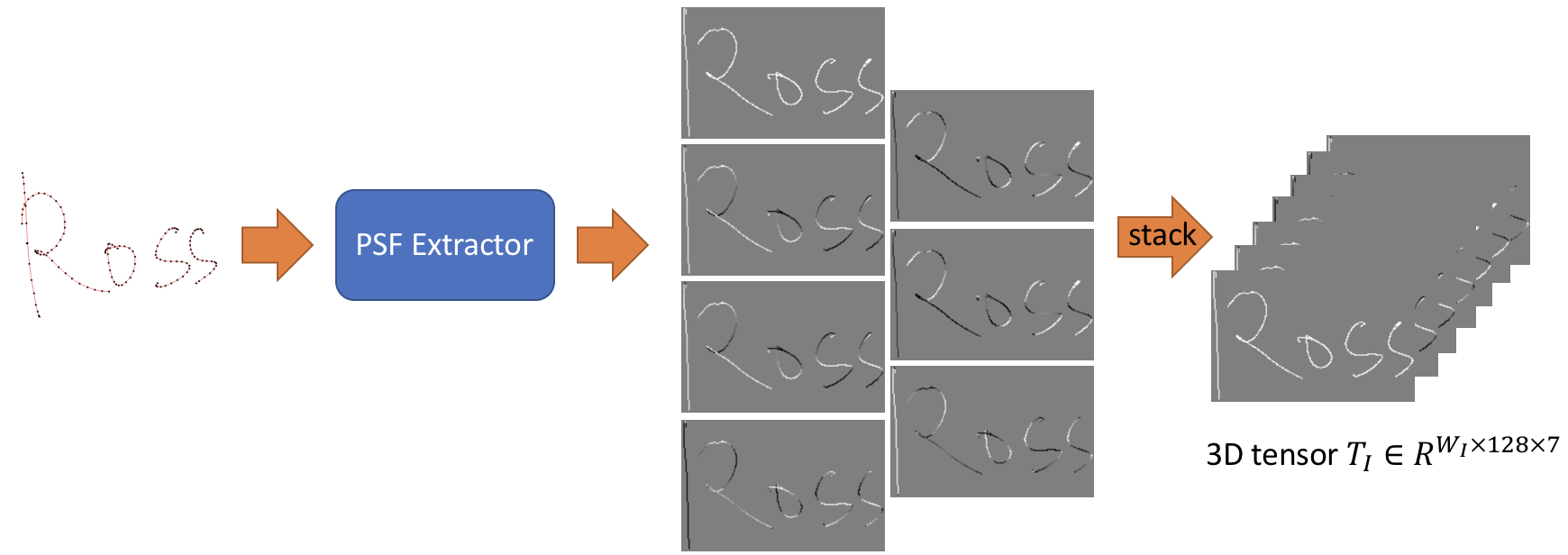}
\caption{Illustration of path signature feature extraction (up to 2nd order), where handwritten text instances are scaled to fixed height as 128, but varying  width denoted as $W_I$.  }
\label{figure:psf}
\end{figure}
These seven path signature maps extracted as described above are stacked to form a seven-channel tensor denoted as $T_I$, which serves as the input of a following CNN-LSTM binary classification model, as illustrated in Figure~\ref{figure:psf}. The CNN-LSTM model consists of a CNN and an LSTM, whose configurations are displayed in Table~\ref{tab:cnn-lstm}. CNN serves to encode extracted path signature feature into a matrix, since the input 3D tensor possesses variable widths, consequently results in that the output matrix of CNN possesses variable columns. Then each column of this encoded matrix is further fed into an LSTM cell sequentially, an FNN is assembled on the top of last output of the LSTM to calculate the probability of current instance as genuine written text. Thus loss function of discriminator is naturally selected as binary cross entropy loss. The whole procedure is shown in Figure~\ref{figure:cnn-lstm}.

\begin{table}[h]
\small
\center
\caption{Architecture Configuration of CNN-LSTM Model, where input shape, kernel shape and output shape are in the manners of (width, height, channel), (\# of kernels, mask, stride) and (width, height, channel) respectively. The stride can either be a single number for both width and height stride, or a tuple (sW, sH) representing width and height stride separately.}
  \begin{tabular}{|c|c|c|c|c|}
  \hline
             & Input Shape   & Kernel Shape          &     Output Shape & Hidden Size  \\ \hline
  Conv1   &  $(W_I, 128, 7)$  & $(32, 3, 1)$        &  $(W_I, 128, 32)$   &   -- \\ \hline
  AvgPool   &  $(W_I, 128, 32)$  & $(-, 2, 2)$        &  $(W_I/2, 64, 32)$   &   -- \\ \hline
  Conv2   &  $(W_I/2, 64, 32)$  & $(64, 3, 1)$        &  $(W_I/2, 64, 64)$   &   -- \\ \hline
  AvgPool   &  $(W_I/2, 64, 64)$  & $(-, 2, 2)$        &  $(W_I/4, 32, 64)$   &   -- \\ \hline
  Conv3   &  $(W_I/4, 32, 64)$  & $(128, 3, 1)$       &  $(W_I/4, 32, 128)$   &   -- \\ \hline
  Conv4   &  $(W_I/4, 32, 128)$  & $(256, 3, (1, 2))$        &  $(W_I/4, 16, 256)$   &   -- \\ \hline
  AvgPool   &  $(W_I/4, 16, 256)$  & $(-, 2, 2)$      &  $(W_I/8, 8, 256)$   &   -- \\ \hline
  Conv5   &  $(W_I/8, 8, 256)$  & $(128, 3, (1, 2))$       &  $(W_I/8, 4, 128)$   &   -- \\ \hline
  Conv6   &  $(W_I/8, 4, 128)$  & $(256, 3, (1, 2))$       &  $(W_I/8, 2, 256)$   &   -- \\ \hline
  AvgPool   &  $(W_I/8, 2, 256)$  & $(-, 2, 2)$       &  $(W_I/16, 1, 256)$   &   -- \\ \hline

  LSTM     & 256 & -- & -- & 256 \\\hline
  FC1      & 256 & -- & --  & 128 \\\hline
  FC2      & 128 & -- & --  & 1 \\\hline  
\end{tabular}
\label{tab:cnn-lstm}
\end{table}

\subsection{Generator}\label{subsec.generator}

\begin{figure}[ht]
    \centering
    \includegraphics[width=1\linewidth]{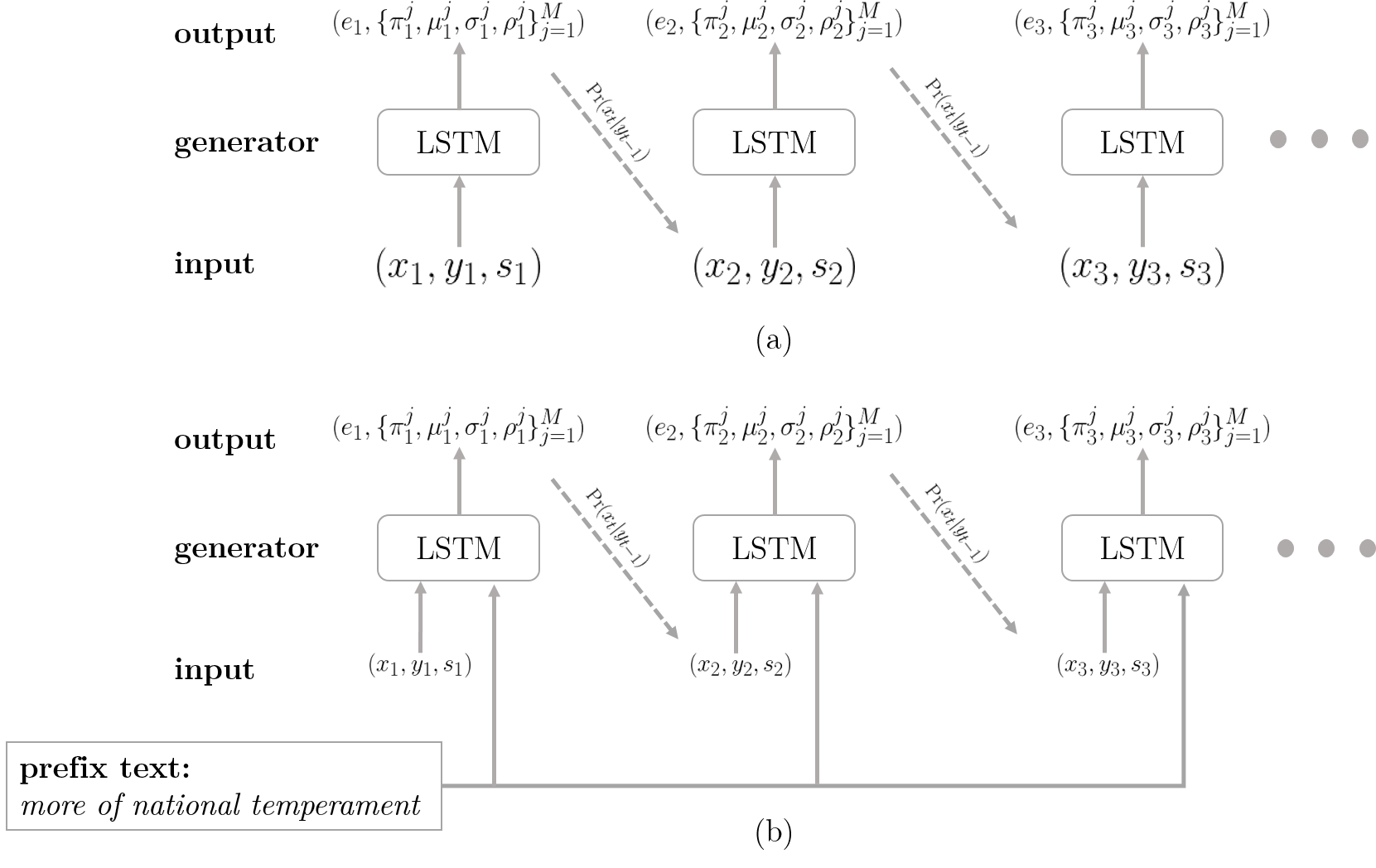}
    \caption{(a) Diagram of handwritten generator for prediction $\mathcal{G}_p$; (b) Diagram of handwritten generator for synthesis $\mathcal{G}_s$; where $M$ is the number of gaussian distributions with parameters as $e, \pi, \mu, \sigma, \rho$.}
    \label{fig:generatordiagram}
\end{figure}

Generator $\mathcal{G}$ is basically stacked LSTMs with sampling layers for generating strokes points from a certain initial stroke point. In this paper, we consider two architectures of generator designed for prediction and synthesis tasks, denoted as $\mathcal{G}_p$ and $\mathcal{G}_s$ proposed by Alex~\cite{graves2013generating}, illustrated as Figure~\ref{fig:generatordiagram}. The prediction generator $\mathcal{G}_p$ only requires the network to generate random handwritten strokes, while the synthesis $\mathcal{G}_s$ generator further asks the model to produce  handwriting texts given prescribed text of user.


Although these two generators are used in different scenarios, they share numerous things in common. First of all, both of them are stacked LSTMs with outputs as parameters of an appropriate probability distribution to predict the status of next stroke point given current generated stroke path. For this probability distribution, Alex~\cite{graves2013generating} proposed to use a mixture of bivariate Gaussian distribution associated with a Bernoulli distribution to represent the distribution of pen up/down. Our model incorporates the same idea, more specifically the outputs of last layer of LSTMs represent the parameters of distribution for x-axis and y-axis offsets between current and next stroke point, and a parameter to indicate if next stroke point ends up stroke path or not.
Secondly, both generators update their parameters by maximizing log-likelihood of generated stroke sequences. During the adversarial training, they utilize the signal from the discriminator described in Section~\ref{subsec.discriminator} to adjust network parameters. Since the discriminator predicts the probability of handwritten text as genuine, the generator aims at maximizing this signal by using generated fake handwritten texts.  Thirdly, both $\mathcal{G}_p$ and $\mathcal{G}_s$ use biased sampling~\cite{graves2013generating} to achieve higher quality of handwriting generation. In terms of the input of discriminator formed as Path Signature Features, the generated handwritten strokes are further processed as PSF features. 



Different from the prediction generator $\mathcal{G}_p$, the inputs of synthesis generator $\mathcal{G}_s$ requires not only the stroke point data but also the text information, since the synthesis network needs to know what texts to generate. The text is an one-hot vector processed from the input string, and input to LSTM layers with window parameters to constrain the synthesis network.

\section{Experiments}\label{sec.experiments}

In this section, we numerically demonstrate the superiority of the adversarial network we proposed than \alexrnn{}.  At first, we describe the experimental settings in Section~\ref{sec:expsetting}, then in Section~\ref{sec:numresults} we move on to presenting numeric results generated from both \alexrnn{} and \algcro{} to illustrate the advantages of our model.

\subsection{Experimental Settings}\label{sec:expsetting}

Both of the prediction network $\mathcal{G}_p$ and synthesis network $\mathcal{G}_s$ are trained on IAM dataset~\cite{marti2002iam}, which contains 80 distinct characters, of which we only choose 54 characters, including 26 capital letters, 26 lower case letters, 1 space character and a place holder for unknown character. Each stroke point of data possesses  x-axis and y-axis offset from the previous stroke point and a binary end-of-stroke indicator.  The source code is available at \url{https://github.com/orris27/hw_gan}.

We use 20 bivariate Gaussian mixture components at the output layer to parameterize the distribution of next point's coordinates and a Bernoulli distribution to determine pen up/down. A mixture of 5 Gaussian distributions was employed for estimating window parameters. The bias for biased sampling is set to 3.0. Architecture configuration details of discriminator $\mathcal{D}$ and two generators $\mathcal{G}_p$, $\mathcal{G}_s$ are shown in Table~\ref{tab:cnn-lstm} to~\ref{tab:generator-synthesis} respectively. The initial learning rates for both generator and discriminator are set as 0.001, then decaying exponentially during training, and optimizer is selected as Adam. For synthesis network $\mathcal{G}_s$, we utilize 25 stroke points to represent one character. In order to ensure that two inputs for the discriminator do not differ significantly in the manner of size during the adversarial training, we resize the PSF features extracted from IAM dataset and those sampled from generator to similar sizes. To avoid exploding gradients, gradients of generator parameters are normalized to magnitude as one, thus step size is fully determined by learning rate, and gradient estimation serves providing an ascent search direction.

It is widely acknowledged that training GANs is relatively difficult~\cite{martin2017gans, salimans2016improved}, and we employ multiple tricks  to train GANs more effectively. Firstly, we normalize PSF features between -1 and 1. Secondly, label smoothing technique is applied to both generator and discriminator during adversarial training~\cite{salimans2016improved}. Besides these, discriminator $\mathcal{D}$ keeps trained with positive samples from IAM dataset and negative samples from generators $\mathcal{G}_s$ and $\mathcal{G}_p$ supervisedly during adversarial training, since a challenge that we face when training the model is to handle the pool capability of discriminator to distinguish realistic handwritten texts generated by $\mathcal{G}_s$ and $\mathcal{G}_p$. More specifically,  in the prediction network, the positive and negative samples fed for discriminator are randomly chosen from IAM dataset and randomly produced by $\mathcal{G}_p$, while in the synthesis network, we first randomly select a set of handwritten text instances with corresponding texts as positive samples, and then generate a set of fake data based on these texts as negative samples. 

\begin{table}[ht]
\small
\center
\caption{Architecture Configuration of LSTM-based Generator for Prediction.}
\begin{tabular}{|c|c|c|c|c|}
\hline
& Input Shape  & Hidden Size \\ \hline
LSTM1 & 3 & 512 \\ \hline
LSTM2& 515 & 256\\ \hline
LSTM3 & 259 & 512 \\ \hline
FC & 1280 & 121 \\ \hline
\end{tabular}
\label{tab:generator-prediction}
\end{table}

\begin{table}[ht]
\small
\center
\caption{Architecture Configuration of LSTM-based Generator for Synthesis.}
\begin{tabular}{|c|c|c|c|c|}
\hline
& Input Shape & Hidden Size \\ \hline
LSTM1 & 57 & 512 \\ \hline
LSTM2& 569 & 512 \\ \hline
FC1 & 512 & 15\\ \hline
FC2 & 512 & 121 \\ \hline
\end{tabular}
\label{tab:generator-synthesis}
\end{table}

\subsection{Numerical Results}\label{sec:numresults}


In this section, experimental results are present to further demonstrate the advantages of our \algcro{} to \alexrnn{}. In general, the \algcro{} can produce competitive handwritten samples in prediction task, more promisingly, it can generate handwritten samples of which stroke points tend to distribute more spatially uniformly than those sampled from \alexrnn{}, and the font styles of handwritten text from \algcro{} are more neat and sometimes more diverse than those of \alexrnn{}. Note that generating more diverse font styles of handwritten text is crucial for many related tasks, e.g. handwritten text with more diverse font styles may result in a more robust handwritten recognition system, which requires a large amount of annotated data to be trained.

\begin{figure}[ht]
    \centering
    \includegraphics[width=.8\linewidth]{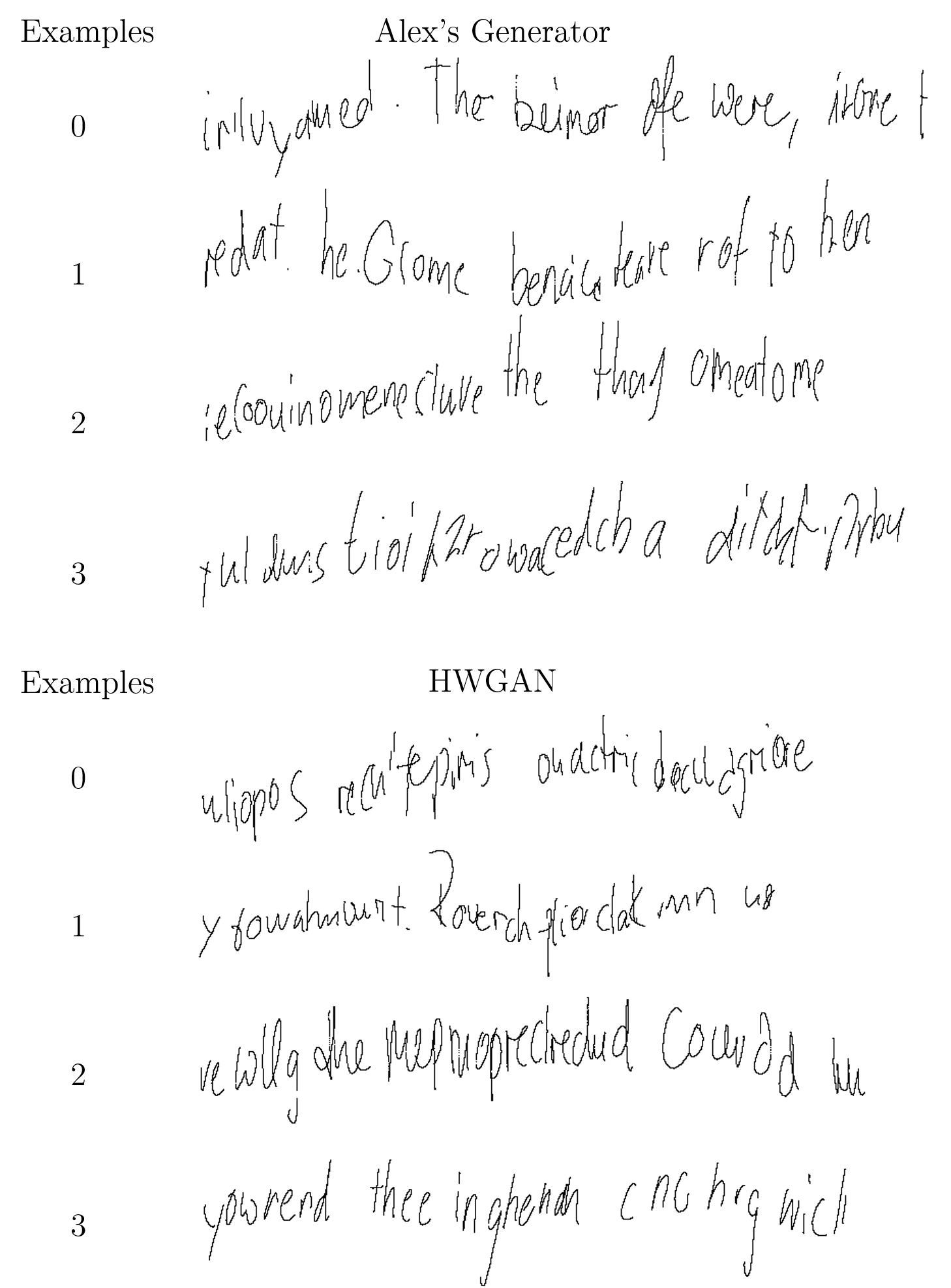}
    \caption{Prediction handwriting samples randomly generated by \alexrnn{} and \algcro{}. The top line in each block is generated by \alexrnn{}, while the others are from \algcro{}.}
    \label{fig:sample_prediction}
\end{figure}

We first describe the performance of our model on prediction tasks compared with \alexrnn{}. The handwriting instances are sampled with bias as 3.0. As shown in Figure~\ref{fig:sample_prediction}, handwriting samples produced by \algcro{} are very competitive to that of Alex, which demonstrates the excellent abilities of learning handwriting structure of GANs. Since the criteria of judging the quality of handwriting data varies from person to person and there is no universal standard, it is not suitable to claim that one is better than the other. However, the samples from two networks are slightly different in font styles, since IAM is the only dataset for training both models, it implies that their learning strategies are not the same and may be applied in different scenarios.

\begin{figure}[h]
    \centering
    \includegraphics[width=.8\linewidth]{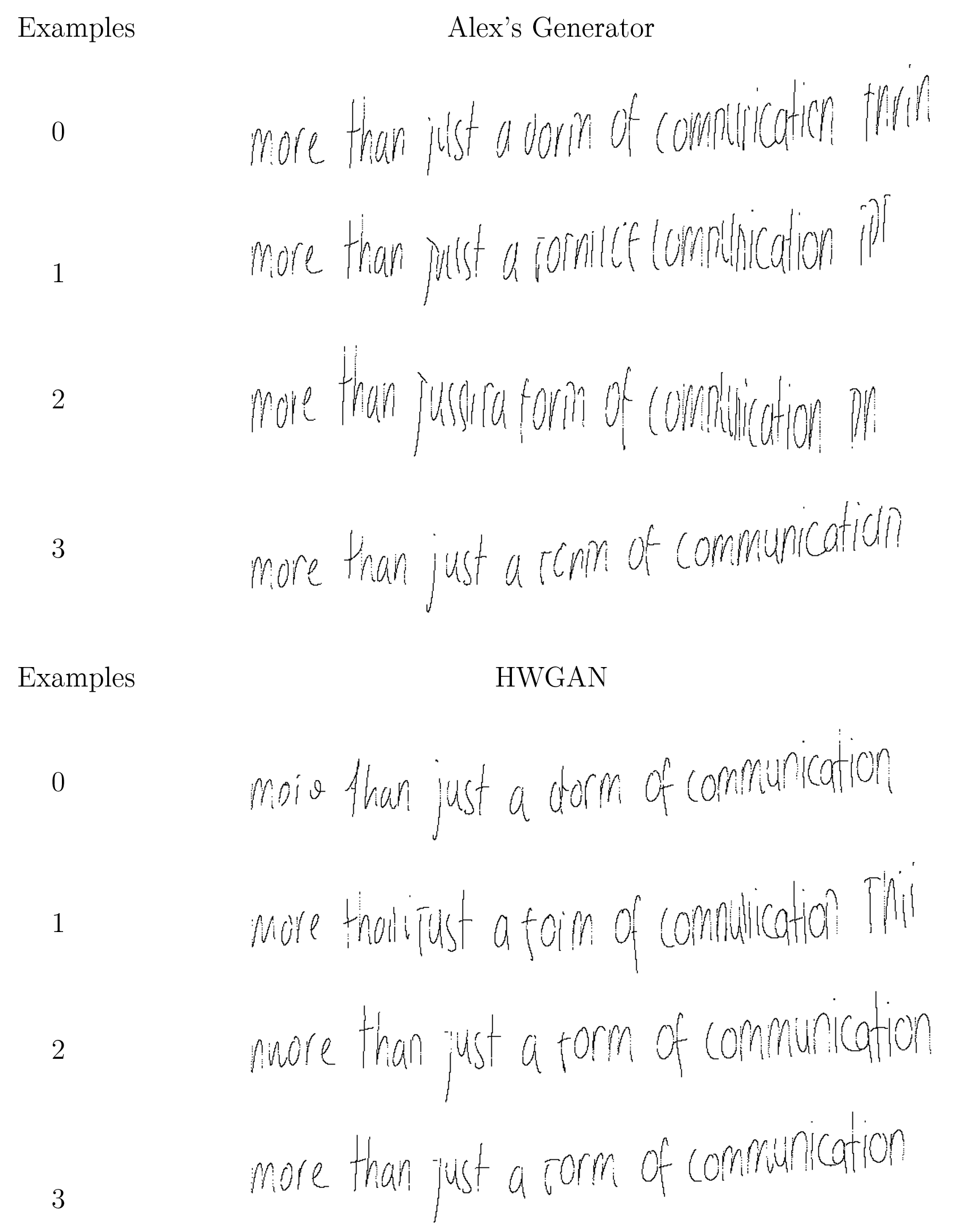}
    \caption{Generated handwritten texts by \alexrnn{} and \algcro{}. The top block is from \alexrnn{}, the bottom one is from \algcro{}. }
    \label{fig:sample_totoal_synthesis}
\end{figure}

We then compare our model with \alexrnn{} in synthesis task. As an overview, the results of both models are shown in Figure~\ref{fig:sample_totoal_synthesis}. Generally speaking, we can see that \algcro{} may produce more realistic, more natural and more neat handwritten texts than \alexrnn{}. More detailed comparisons are followed in the remaining section.

\begin{figure}[ht]
    \centering
    \includegraphics[width=.7\linewidth]{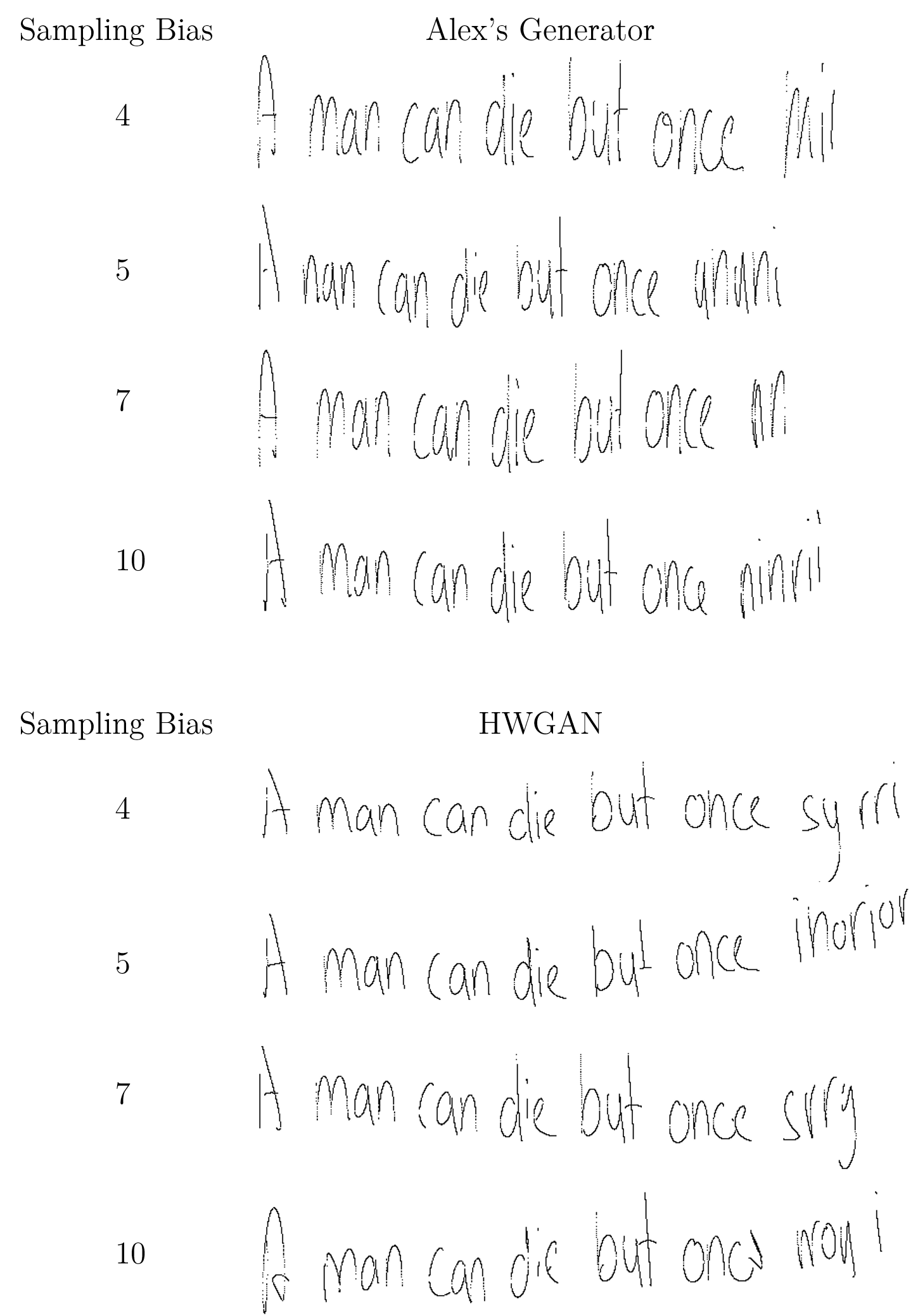}
    \caption{Synthesis handwriting data from \alexrnn{} and \algcro{} using different sampling bias. The top block is sampled from \alexrnn{}, the bottom one is from \algcro{}.}
    \label{fig:sample_sampling_bias}
\end{figure}

Given that the value of bias used in biased sampling has a large impact on the results, we next test both models under different sampling biases. In Figure~\ref{fig:sample_sampling_bias}, the top handwritten samples are from \alexrnn{}, and the bottom samples are from \algcro{} under sampling biases as 4, 5, 7 and 10 respectively. The results of \algcro{} distribute almost more spatially uniformly, and neat on every bias. On the other hand, it seems that \algcro{} may produce more diverse handwritten text styles, like the ``A".

\begin{figure}[h]
    \centering
    \includegraphics[width=.75\linewidth]{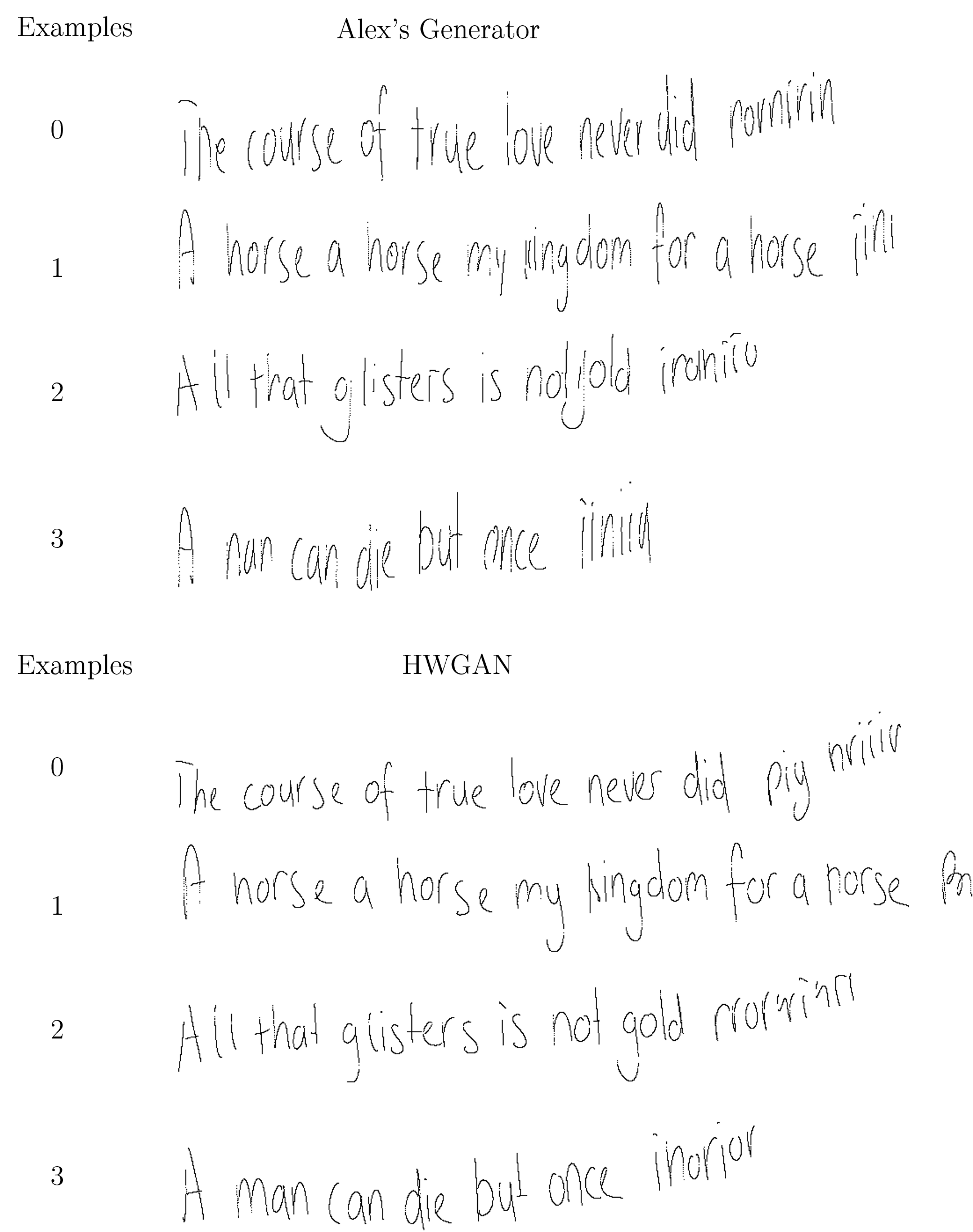}
    \caption{Synthesis handwriting data from \alexrnn{} and \algcro{} given different characters. The top block is sampled from \alexrnn{}, and the bottom block is generated from \algcro{}.}
    \label{fig:sample_characters}
\end{figure}

Besides the above, we also investigate both models based on different prescribed sentences with a fixed sampling bias as 5.0, and present the results in Figure~\ref{fig:sample_characters}, which further proves the advantages of our model observed on previous experiments are consistent.

\section{conclusions}\label{sec.concolusion}

We present a new Generative Adversarial Network architecture \algcro{} for synthesizing handwritten text of digital ink.  The method comprises two main components, a discriminator consisting a PSF feature extractor, followed by a CNN-LSTM binary classifier to distinguish realistic and forgery handwritten data, and a generator following the architectures of Alex Grave's to produce either random handwritten characters or sentences with prefixed text. In particular, the numerical results show that \algcro{} are generally better than \alexrnn{} in terms of uniformness of spatial distribution of stroke points, neatness and diversity of font style. This phenomenon illustrates that adversarial training is beneficial for generating more realistic handwritten text data. 

\bibliographystyle{unsrt}
\bibliography{bibliography}

\end{document}